\pdfoutput=1

\documentclass[11pt]{article}

\usepackage{acl}

\usepackage{times}
\usepackage{booktabs}
\usepackage{latexsym}
\usepackage{subcaption}
\usepackage{graphicx}
\usepackage{multirow}
\usepackage{subcaption}
\usepackage{booktabs}
\usepackage{pifont}
\usepackage{xcolor,colortbl}
\usepackage{pgfplots}
\usepackage{tikz}
\usepackage{xspace}

\usepackage[T1]{fontenc}

\usepackage[utf8]{inputenc}
\usepackage{amsmath}
\usepackage{empheq}
\usepackage{MnSymbol}
\usepackage{microtype}

\usepackage{graphicx}

\title{ Communicating with  Speakers and Listeners of Different Pragmatic Levels }
 \author{Kata Naszádi\textsuperscript{1}
 \and Frans A. Oliehoek \textsuperscript{2}
 \and Christof Monz\textsuperscript{1} \\
 \textsuperscript{1}Language Technology Lab, University of Amsterdam \\
 \textsuperscript{2}Delft University of Technology\\
{\fontfamily{qcr}\selectfont \small \href{mailto:k.naszadi@uva.nl}{k.naszadi@uva.nl}  } \\ }

\begin{document}
\maketitle
\begin{abstract}
This paper explores the impact of variable pragmatic competence on communicative success through simulating language learning and conversing between speakers and listeners with different levels of reasoning abilities. Through studying this interaction, we hypothesize that matching levels of reasoning between communication partners would create a more beneficial environment for communicative success and language learning. Our research findings indicate that learning from more explicit, literal language is advantageous, irrespective of the learner's level of pragmatic competence. Furthermore, we find that integrating pragmatic reasoning during language learning, not just during evaluation, significantly enhances overall communication performance. This paper provides key insights into the importance of aligning reasoning levels and incorporating pragmatic reasoning in optimizing communicative interactions.

\end{abstract}

\section{Introduction}
In everyday conversations there is a trade-off between clarity and conciseness. Efficient messages might appear under-specified or ambiguous under a literal interpretation but can be successfully resolved using pragmatic reasoning about the speaker's intentions and the context of the communication \citep{grice1975logic, horn1984meaning, fox2011characterization, davies2022speaker}. If the speaker trusts the listener to make the right inferences, they can choose to be more concise.  Being able to infer the intended meaning of an utterance beyond its literal content allows us to communicate efficiently. 

The process of how people attain pragmatic interpretations using a model of the speaker's intentions has long been studied. There is also plenty of evidence from psycho-linguistic studies that individuals have different levels of pragmatic competence \cite{franke2016reasoning, mayn2023individual}. More importantly, people have been shown to keep track of the communicative partner's pragmatic competence and adjust their interpretations and messaging accordingly. This has been demonstrated both with human \cite{horton2002speakers, mayn2024beliefs} and artificial partners \citep{loy2023perspective, branigan2011role}.

\begin{figure}
    \centering
    \includegraphics[scale=0.45]{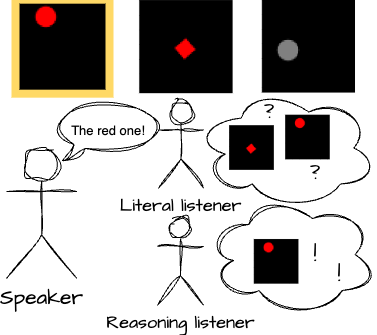}
    \caption{The speaker is asking for the red object. For a literal listener, this is ambiguous. A reasoning
listener considers alternative messages about shape and color features and concludes that the speaker is asking for the red circle, as ”square" would have been a more informative message
for the other red object. }
    \label{fig:red-example}
\end{figure}

The pragmatic reasoning modeled in this work involves counterfactual reasoning about alternative sentences that the speaker could have uttered . The interaction in Figure \ref{fig:red-example} depicts an instance of such pragmatic reasoning about alternatives within our simple environment. According to pragmatic theory \citep{grice1975logic} the same process accounts for the interpretation "They are in the office for the rest of the week", when we hear the sentence "We are not in the office on Mondays".

In this work, we investigate the impact of varying pragmatic competence on communicative success. We pair literal and pragmatic listeners with speakers of different levels of pragmatic competence. We study the interaction between such speakers and listeners not only during inference, where both partners have an already learned lexicon, but also during language learning. This way we gain insight into optimal levels of pragmatic inference for teachers and language learners. We hypothesise that matching levels of reasoning between partners benefits communicative success and language learning. 

Our simulations reveal that with a lexicon that doesn't perfectly match that of the speaker's, sophisticated pragmatic listeners still significantly benefit from explicit literal language use. We also show that language learners that do not model pragmatic inference, struggle when learning from a speaker who uses pragmatic communication, while language learners that integrate a model of the speaker are significantly more successful.

\section{Background}
We situate our listener in an image-based version of Lewis's signaling game \cite{Lewis1969-LEWCAP-4}. Image-referential games are commonly  used to study the benefit of speakers and listeners reasoning about each other in context \citep{lee2018answerer, white2020learning, andreas2016reasoning}.  

At each turn a collection of N images is provided as context $C=(o_{1},...,o_{N})$, with the speaker having knowledge of a specific target image $o_{t}$, where $1\le t \le N$. The listener's objective is to correctly identify the target image index $t$ given the speaker's message $w$ . The messages may contain multiple words by combining words from a fixed vocabulary.
\subsection{Literal meanings and the Rational Speech Act model}
\label{subsec:rsa}

 \citet{goodman2012} provide a concise  model for how speakers and listeners reason about each-other when sharing referential content. As a starting point, the model assumes an underlying literal interpretation. This is a function $D(w,o)$ of an utterance $w$ and an  observation $o$, in our case an image. In the original formulation the base interpretation function is a 0-1 valued indicator of the set of messages that are true of the image $o$. In line with other work, we replace this binary function with a real-valued similarity between the observed image-embedding and text-embedding.
\begin{equation}
\label{eq:embedding}
   D(o_i, w) = \mathrm{CNN}_{\theta} (o_i)^T \mathrm{RNN}_{\theta}(w)
\end{equation}
 Each image $o_i$ is individually embedded with a CNN following the ResNet architecture \cite{He_2016_CVPR}. The embedding  if the message $w$ is computed by an RNN with Gated Recurrent Units \cite{cho-etal-2014-learning}.

The listener models the distribution over the indices in an ordered set of images. The simplest listener distribution is produced by normalizing the score assigned by literal interpretation function over all the images in a given context $C$.
\begin{equation}
    L_0(i|w,C)=  \frac{ e^{ D(o_i,w)}}{\sum_{j=1}^{|C|} e^{D(o_j,w)}}
\end{equation}

The speaker produces a message that maximizes the probability that the listener chooses the right image and also considers the cost of each message $w$. This means that the speaker has an internal model of the listener.
\begin{equation}
\label{s1}
   S_n(w|C,i) = \frac{e^{\lambda(\log (L_{n-1}(i|C,w)) - \mathrm{cost}(w))}}{\sum_{w' \in V} e^{  \lambda( \log(L_{n-1}(i|C,w')) - \mathrm{cost}(w'))}}
\end{equation}

In this work, we use a cost function that
assigns a constant weight to each word and we only consider fully rational speakers with $\lambda =1$. In the case of the speaker, the normalization happens over all possible  messages $w \in V$. 
This is the most expensive step in the hierarchical reasoning process. In many natural language applications it is even prohibited by the fact that the set of all possible utterances is infinite. While exact inference is intractable, there are many papers discussing approximations \citep{Cohn-GordonGP18, liu2023computational, lazaridou2020multi, white2020learning}. In our communication-game, messages may contain one or two words: naming either the shape or the color of the target or both.

Building on \ref{s1}, higher level listeners have an internal model of a speaker:
\begin{equation}
\label{L_2}
    L_n (i | C, w ) \propto S_{n-1}(w|C,i)  P(C,i)
\end{equation}
By applying Equations \ref{s1} and \ref{L_2} in an alternating fashion, we can produce higher level speakers and listeners. 

The most studied levels in the case of human communication are $L_0$ literal and $L_2$ pragmatic listeners paired with $S_1$ and $S_3$ speakers. This is motivated by evidence that humans can interpret messages from a $S_3$ speaker consistent with a $L_2$ listener \cite{goodman2016pragmatic} and multiple pragmatic phenomona have been derived using the RSA framing and these levels \citep{franke2016reasoning, hawkins2023partners}.
\subsection{Reasoning while learning}
\label{sec:model-desc}
In the previous subsection \ref{subsec:rsa} we saw how to perform recursive reasoning on top of given literal representations $D(o,w)$.  These literal interpretations are most commonly initialized by functions learned outside of the context of a referential game and the reasoning is added only during inference \citep{fried-etal-2018-unified, lazaridou2020multi, andreas2016reasoning, liu2023computational}.  

However, the optimal literal representations are likely influenced by the reasoning itself. Following the work of \citet{monroe2015learning} and \citet{mcdowell-goodman-2019-learning}, we would like to integrate the knowledge that the received messages are the result of pragmatic reasoning already during learning. Therefore, we apply recursive reasoning during model training. 

Pragmatic listeners seek to update the weights of the literal interpretation $D(o,w)$ but they need to do so by considering the repeated application of Equations~\ref{s1}  and \ref{L_2}. Similarly to \citet{mcdowell-goodman-2019-learning}, we derive the gradients of the reasoning process with respect to the lexicon weights. By repeated application of the chain rule through the hierarchical reasoning, pragmatic listeners  backpropagate through the hierarchical reasoning and update the weights of the image- and  utterance-embedding models.

\section{Data}
\label{data}

To investigate the impact of the pragmatic competence of speakers and listeners on communicative success, it is necessary to establish a controlled setting that allows for manipulation of the reasoning abilities of participants. We create a new environment based on the ShapeWorld dataset \cite{Kuhnle2017ShapeWorldA}. Instead of the rule based method of \citet{Kuhnle2017ShapeWorldA}, we use an exact implementation of the rational speaker defined in Equation \ref{s1}. This way we can create speakers with different depth of recursive reasoning. Our speakers are not learned, they are knowledgeable users of the language: they have access to the underlying true lexicon which indicates the mapping between color and shape words and image properties.

Each game consists of a target image and a variable number of $N-1$ distractor images. Images are described by one out of six different colors and a shape that can take five different values. 
The location, size and rotation of the objects is randomized on a 64x64 grid which creates a large variation of candidate pictures. 

 We parameterize the process that generates the image tuples for each game by four probability distributions: the priors over the shapes $P(S)$ and colors $P(C)$, the probability that controls the correlations between colors $P(C|C)$ and the conditional defining the co-occurrence of shapes $P(S|S)$. We sample these distributions from different Dirichlet-distributions. We create two sets of concentration parameters: in the first version of the game, all sampled distributions are close to uniform ($Corr=0$), while in the second version introduces correlations in the shape and color conditionals ($Corr=1$). This way the sampled image tuples share more features, creating higher likelihood for pragmatic messaging that differentiates $S_1$ and $S_3$.  

For training, we sample only one instance of each distribution. At test time, we sample different $P(S)$, $P(C)$, $P(S|S)$ and $P(C|C)$ instances 10 times. From each of these constellations we sample 3200 games. 

The random seed is fixed across all experiments and is reset for the learning and evaluation of each learner. This ensures that each listener sees the exact same examples in all environments.

\section{Experiments}

The fact that we have full control over the speaker's messaging strategy and the data generating process allows us to alter the level of the speakers that the listeners learn from and create image tuples that highlight the contrast between higher level pragmatic and lower level literal messaging strategies.

We train train $L_0$ literal listeners and  $L_2$ pragmatic listeners. We create two  different levels of speakers to pair them with our learning listeners: $S_1$ has an internal model of a competent $L_0$, while $S_3$ anticipates $L_2$-behavior. 

Implementation for training and evaluating all models can be found at \href{https://github.com/naszka/rsa_backward/}{https://github.com/naszka/rsa\_backward/}.

\subsection{Results}

In this section, we present the insights gained from simulating language learning and communication between listeners and speakers with pragmatic or literal preferences. First we look at altering speaker and listener levels only during evaluation using an already trained lexicon. Then we turn to the learning dynamics between our four pairs: $L_0$ - $S_1$, $L_0$ - $S_3$, $L_2$ - $S_1$ and $L_2$ - $S_3$.

\begin{table}[]
    \centering
    \begin{tabular}{ |c||c|c| }

 \hline
 Distractors & $S_1$ & $S_3$ \\
 \hline
 2   & 1.07   & 1.01 \\
 3   & 1.14  & 1.02  \\
 4   & 1.24   & 1.09  \\
 \hline
\end{tabular}
    
    \caption{Average message length in words over 5000 samples for different number of distractors and speaker  levels,  $Corr=1$. Higher level speakers send shorter messages and more distractors result in longer messages.}
    \label{tab:msg_len}
\end{table}

\paragraph{Listening to speakers with different depth}

\begin{table}[!htbp]
\centering
\begin{tabular}{lccl}
\toprule
{} & {Listener eval} & {Speaker eval} & {Accuracy} \\
\midrule
a) &  0 & 3& $80.5$ \\
b) &  2   & 3& $81.2$ ** \\
\hline
c) &0  & 1 &  $85.5$ \\
d) &  2 & 1 &$85.6$ \\
\bottomrule
\end{tabular}
\caption{ A listener trained as $L_0$ upgraded to different listener levels and paired with $S_1$ or $S_3$  at evaluation. Both $L_0$ and $L_2$ perform  significantly  better with the more verbose $S_1$. When receiving messages from an $S_3$, the higher level $L_2$ is significantly better. Evaluation setup: $cost = 0.6$,  $N=5$, $Corr=1$. }
\label{tab:speakerlevels_eval}
\end{table}
First we take the $L_0$ listener which learned in the easiest environment ($S_1$, $Corr=0$, $N=3$) hence has the highest in-domain performance of $91.2 \%$ accuracy. During evaluation, we upgrade this listener to different levels: this means that during inference we apply recursive reasoning on top of the already learned $L_0$ lexicon. We pair these listeners with $S_1$ and $S_3$. Table \ref{tab:speakerlevels_eval}  shows that pragmatic $L_2$ is significantly \footnote{ We perform Fisher's exact test for significance testing. We note $p<0.05$ with one asterisk * and for $p<0.01$ we put ** next to the results.} better than literal $L_0$ when paired with $S_3$. At the same time, $L_2$ still achieves the best performance with the more verbose $S_1$, this is due to the fact that the listener did not learn the word-feature mapping with perfect accuracy and they still benefit from the more descriptive input.

We picked the evaluation parameters shown in Table \ref{tab:speakerlevels_eval} to maximize  the speaker-type effect. The same trends hold for different number of distractors. 
\paragraph{Learning from speakers with different depth}
\begin{table}[!htbp]
\centering
\begin{tabular}{lccl}
\toprule
{} & {Listener} & {Speaker train} & {Accuracy} \\
\midrule
a) & \multirow{2}{*}{0} &  1 &   $80.7$** \\
b) &  &  3 &  $79.1$ \\
\hline
c) & \multirow{2}{*}{2}  & 1  &   $84.8$** \\
d) &  & 3 &   $83.2$ \\

\bottomrule
\end{tabular}
\caption{For each level of listener, learning from  lower level $S_1$ results in significantly better accuracy. Listener levels are kept the same during evaluation and training.
Training  and evaluation setup: $cost = 0.6$,  $N=5$, $Corr=1$. Evaluation: $S_1$.}
\label{tab:speakerlevels_train}
\end{table}
Now we turn to how listeners of different levels are impacted by learning from different speakers. 

Table~\ref{tab:speakerlevels_train} shows that reasoning learners that learned from lower level speakers always achieve higher accuracy at evaluation. 
This can be explained by the fact that lower level speakers send longer messages on average, see Table~\ref{tab:msg_len}, because their internal model is of a simpler listener who needs longer descriptions for success. 
 \begin{figure}[!htbp]
     \centering
     \includegraphics[scale=0.36]{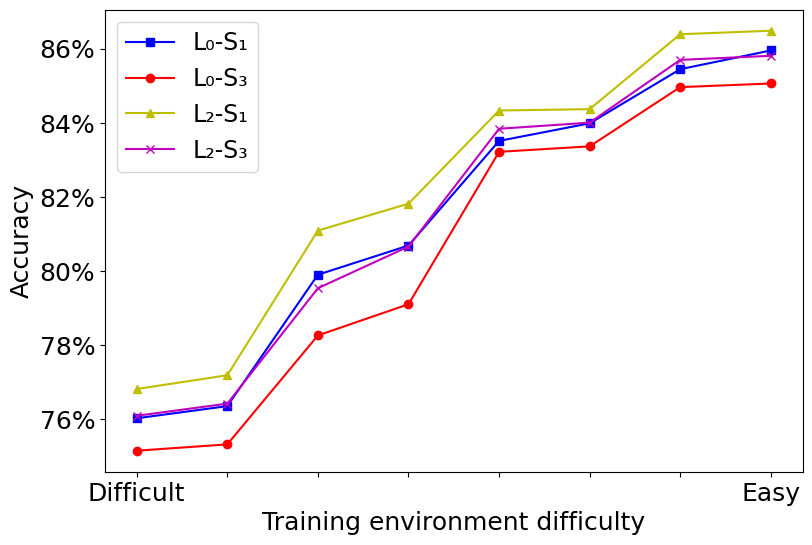}
     \caption{During training, listeners are paired with speakers of different pragmatic competence. The listeners are trained in environments of increasing difficulty. $L_0$ learners paired with $S_1$ speakers have the same performance as $L_2$ paired with $S_3$.}
     \label{fig:spekaer_listener_pairs}
 \end{figure}

 Despite the fact that a $L_2$ can disambiguate $S_3$ messages, learning from  a $S_1$ speaker is easier as it provides more data on both image features. This behaviour nicely aligns with the intuition that language learners benefit from simple, verbose communication and teachers should not assume challenging patterns of communicative competence early on in the learning process \cite{nguyen20228}. 
 
 Comparing all possible pairings in  Figure \ref{fig:spekaer_listener_pairs} however,  we can clearly see the benefit of listeners having the appropriate level for the speaker during learning. A $L_0$ listener learning from a $S_1$ matches the performance of a $L_2$ listener learning from a $S_3$ speaker.  We evaluate listeners that were paired with higher or lower level speakers during training. The evaluation environment is kept the same,  all listeners are upgraded to $L_2$ and deployed with $S_1$. Pragmatic $L_2$ listener can compensate for the difficulty of learning from the concise $S_3$ through all training environments.

\section{Conclusions}

%
%
%
Humans exploit pragmatic reasoning in order to reduce the effort of speaking. For artificial agents to understand humans, it is critical to correctly resolve ambiguities. By recursively modeling the conversational partner, pragmatic listeners can arrive at the interpretations intended by pragmatic speakers.

In this work, we introduced speaker-listener pairs with matching or misaligned levels of pragmatic competence.  We examined the benefits of integrating pragmatics not only during evaluation but already during language learning. Our results show that  learning  from more explicit, literal language is always beneficial, regardless of the pragmatic capacity of the learner. At the same time, we conclude that language learners need to apply reasoning about the context and the speaker when learning from data that was generated pragmatically.

\section{Limitations}
While the conversational phenomena we model in this paper have been widely attested to in linguistic theory and psycho-linguistic research, our experiments are limited to an artificial sandbox scenario with a small vocabulary and simple observations. 
The reasoning about all possible utterances used in this paper is intractable with larger vocabularies.

Real world conversations contain a wide range pragmatic inferences, not all of which can be accounted for by the recursive reasoning presented in this paper.

\section{Acknowledgements}
This research was funded in part by the Netherlands Organization for Scientific Research (NWO) under project number VI.C.192.080. We also received funding from the Hybrid Intelligence Center, a 10-year programme
funded by the Dutch Ministry of Education, Culture and Science through the Netherlands Organisation for Scientific Research with grant number 024.004.022.

\bibliography{short}

\appendix

\section{Model training and implementation}
All 261838 model-parameters are trained from scratch. The weights are updated with the AdamW optimizer \cite{Loshchilov2017DecoupledWD} which we initialize with a learning rate of $1e-5$.

For each training step, we use a batch of 32 games and the listeners are trained for 25920 training steps. Each instance of a listener training took 1.5 GPU hours on a single NVIDIA RTX A6000 GPU.
\section{Concentration parameters of the image generators}

We sample $P(S)$, $P(C)$, $P(C|C)$ and $P(S|S)$ from Dirichlet distributions. In the case of no correlation between the images ($Corr=0$), we set all concentration parameters to 1. For the correlated case ($Corr=1$), we introduce correlation between the same shapes and a randomly chosen shape from all five shapes. We achieve this by setting the concentration parameter $\alpha$ to $5$ at the index that corresponds to the i'th shape and a randomly generated other index. $P(S| S=shape_i) \sim Dir(\alpha_1, ... ,\alpha_5)$, where all $\alpha$'s are $1$ except for $\alpha_i = 5$ and $\alpha_j=5$ for a randomly generated $j$. We apply the same process for generating all the $P(C|C)$ distributions.

\section{Benefits of pragmatic reasoning during learning}
\subsection{Pragmatic listeners learn faster}
\begin{figure}[!htbp]
    \centering
    \includegraphics[scale=0.055]{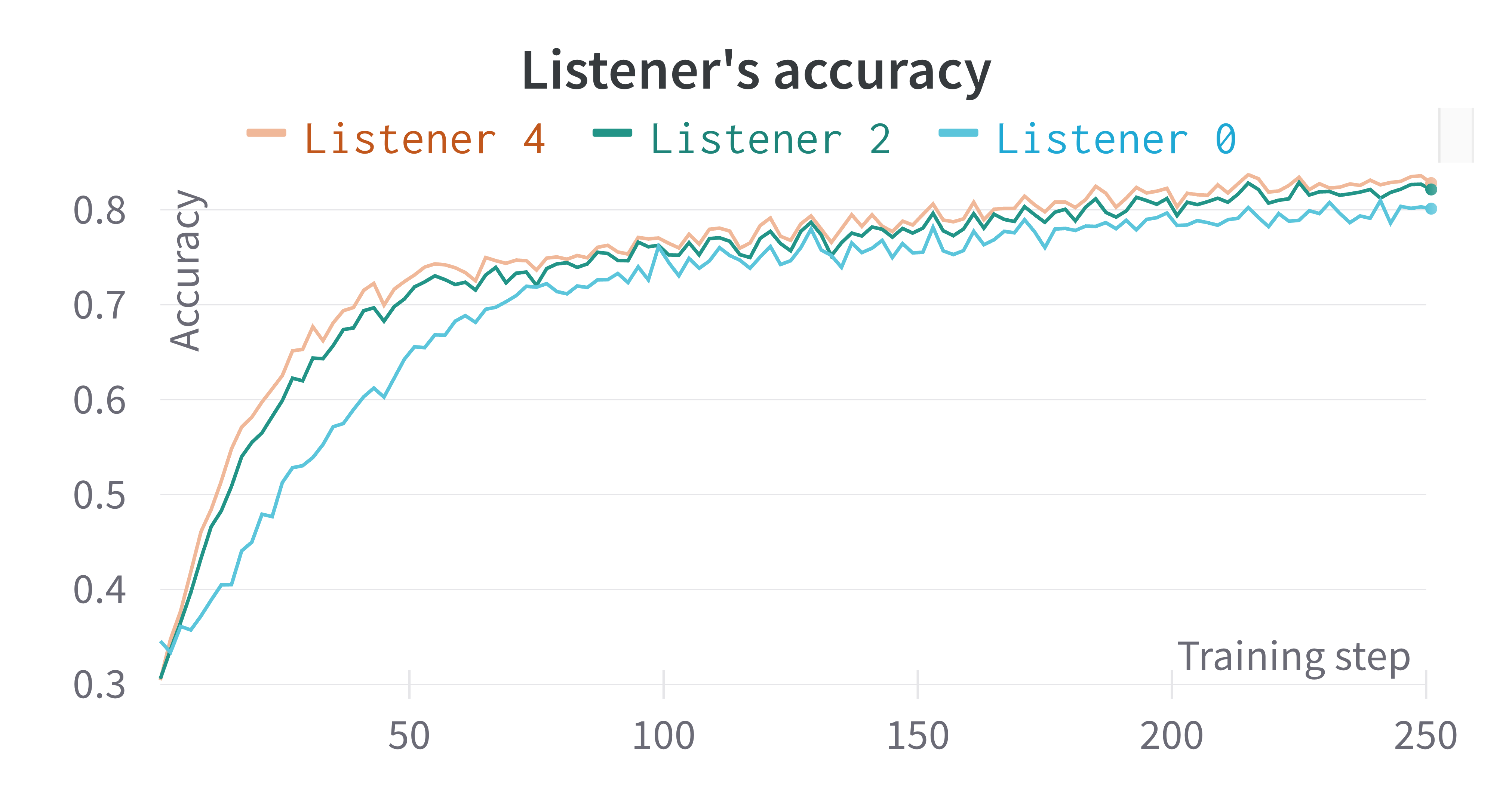}
    \caption{Higher level listeners learn quicker. In this comparison all other parameters such as speaker level, number of distractors, correlation between shapes are left constant. }
    \label{fig:learning}
\end{figure}

Figure~\ref{fig:learning} shows that when we keep all parameters of the learning environment constant, and only vary the listener's depth, we observe that listeners with higher levels, learn to perform the task with good accuracy faster. 
The gap in performance is especially large in the initial learning stages. 
This result is in line with \citet{mcdowell-goodman-2019-learning}, where they discuss the benefits of pragmatic training.

\end{document}